\documentclass[10pt,twocolumn,letterpaper]{article}

\usepackage{cvpr}
\usepackage{times}
\usepackage{epsfig}
\usepackage{graphicx}
\usepackage{amsmath}
\usepackage{amssymb}

\usepackage{tabulary}
\usepackage{multirow}

\usepackage{color}

\usepackage{wrapfig}
\usepackage{enumitem}
\usepackage{adjustbox}

\definecolor{citecolor}{RGB}{34,139,34}
\usepackage[pagebackref=true,breaklinks=true,letterpaper=true,colorlinks,citecolor=citecolor,bookmarks=false]{hyperref}

 \cvprfinalcopy 

\newcommand{\app}{\raise.17ex\hbox{$\scriptstyle\sim$}}

\makeatletter\renewcommand\paragraph{\@startsection{paragraph}{4}{\z@}
  {.5em \@plus1ex \@minus.2ex}{-.5em}{\normalfont\normalsize\bfseries}}\makeatother

\newlength\savewidth

\makeatletter\renewcommand\paragraph{\@startsection{paragraph}{4}{\z@}
  {.5em \@plus1ex \@minus.2ex}{-.5em}{\normalfont\normalsize\bfseries}}\makeatother

\def\cvprPaperID{***} 

\begin{document}

\title{SIXray: A Large-scale Security Inspection X-ray Benchmark\\for Prohibited Item Discovery in Overlapping Images}

\author{Caijing Miao$^{\dag}$, Lingxi Xie$^{\ddag}$, Fang Wan$^{\dag}$, Chi Su$^{\sharp}$, Hongye Liu$^{\sharp}$, Jianbin Jiao$^{\dag}$, Qixiang Ye$^{\dag}\thanks{Corresponding Author}$\\
$^{\dag}$University of Chinese Academy of Sciences \quad $^{\ddag}$The Johns Hopkins University \quad $^{\sharp}$Kingsoft \\
{\tt\small \{miaocaijing16,wanfang13\}@mails.ucas.ac.cn}\qquad{\tt\small 198808xc@gmail.com}\\{\tt\small \{jiaojb,qxye\}@ucas.ac.cn}\qquad{\tt\small \{suchi,liuhongye\}@kingsoft.com}
}

\maketitle

\begin{abstract}
In this paper, we present a large-scale dataset and establish a baseline for prohibited item discovery in Security Inspection X-ray images. Our dataset, named SIXray, consists of 1,059,231 X-ray images, in which 6 classes of 8,929 prohibited items are manually annotated. It raises a brand new challenge of overlapping image data, meanwhile shares the same properties with existing datasets, including complex yet meaningless contexts and class imbalance. We propose an approach named class-balanced hierarchical refinement (CHR) to deal with these difficulties. CHR assumes that each input image is sampled from a mixture distribution, and that deep networks require an iterative process to infer image contents accurately. To accelerate, we insert reversed connections to different network backbones, delivering high-level visual cues to assist mid-level features. In addition, a class-balanced loss function is designed to maximally alleviate the noise introduced by easy negative samples. We evaluate CHR on SIXray with different ratios of positive/negative samples\footnote{Throughout this paper, images with at least one prohibited item are called ``positive'', otherwise called ``negative''.}. Compared to the baselines, CHR enjoys a better ability of discriminating objects especially using mid-level features, which offers the possibility of using a weakly-supervised approach towards accurate object localization. In particular, the advantage of CHR is more significant in the scenarios with fewer positive training samples, which demonstrates its potential application in real-world security inspection.

\end{abstract}

\section{Introduction}
\label{sec_introduction}

\begin{figure}[t]
\centering
\textcolor{red}{Dataset: {\tt\small https://github.com/MeioJane/SIXray}}\\
\vspace{0.2cm}
\includegraphics[width=\linewidth]{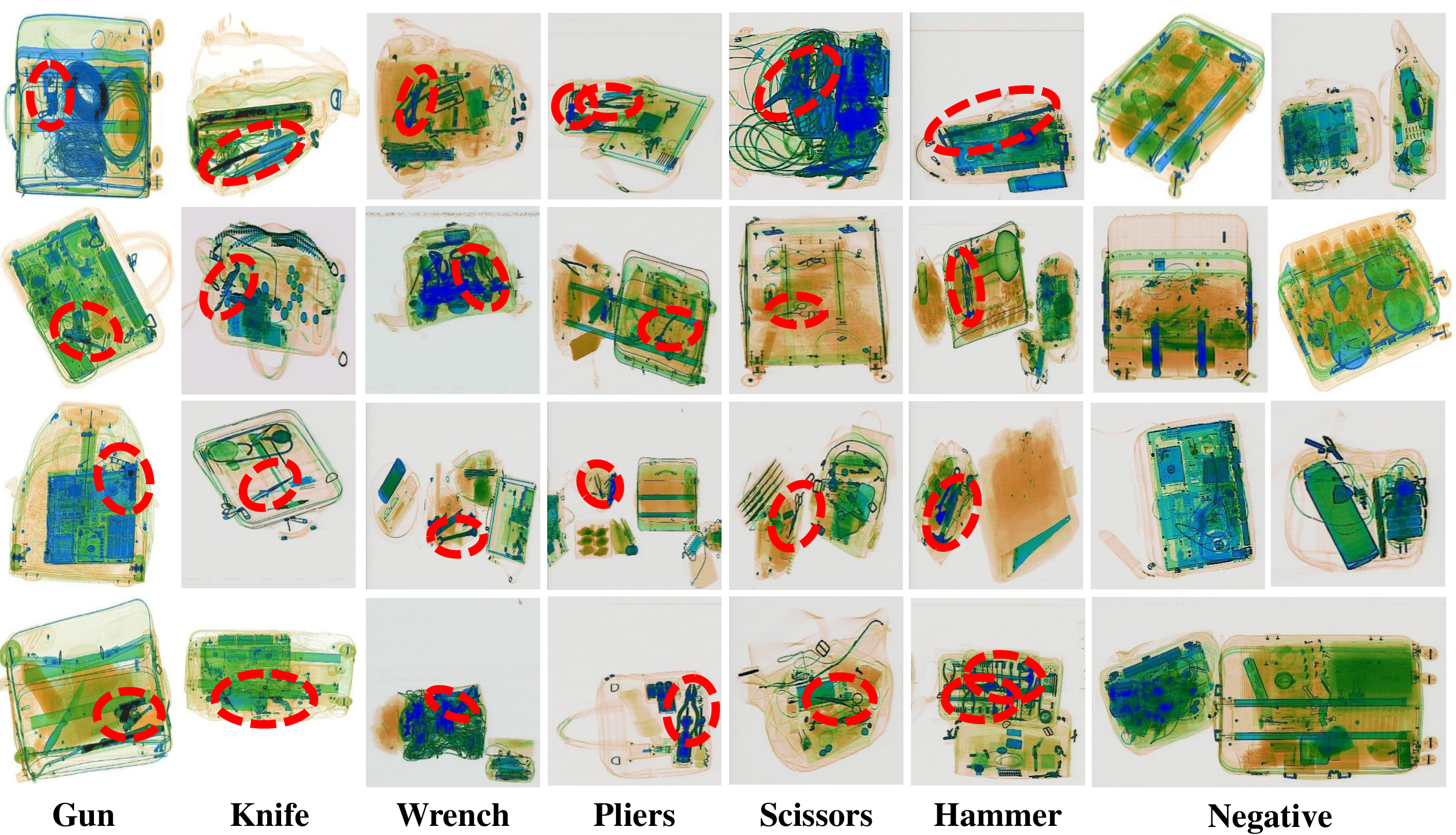}
\caption{Example images in the presented SIXray dataset with six categories of prohibited items. Challenges include large variety in object scale and viewpoint, object overlapping and complex backgrounds (please zoom in for details).}

\label{fig:dataset_examples}
\end{figure}

Security inspection has been playing a critical role in protecting public space from safety threatening such as terrorism. With the growth of population in large cities and crowd density in public transportation hubs, it becomes more and more important to fast, automatically and accurately recognize prohibited items in X-ray scanned images. Recent years, the rapid development of deep learning~\cite{lecun2015deep} in particular convolutional neural networks has brought an evolution to image processing and visual understanding, including discovering and recognizing objects in X-ray images~\cite{mery2013x}\cite{mery2017modern}\cite{mery2017automatic}. Different from natural images and other X-ray scans~\cite{wang2017chestx}, security inspection often deals with a baggage or suitcase where objects are randomly stacked and heavily overlapped with each other. Therefore, in the scanned images, the objects of interest may be mixed with arbitrary and meaningless clutters and thus can be ignored even by human inspectors, Figure~\ref{fig:dataset_examples}.

To provide a public benchmark for research in this field, in this paper, we present a dataset named Security Inspection X-ray (SIXray), which is $100$ times larger than the existing largest image collection for prohibited item discovery, {\em i.e.}, the {\em baggage} group in the GDXray dataset~\cite{mery2015gdxray}. SIXray contains more than one million X-ray images in which only less than $1\%$ images have positive labels ({\em i.e.}, prohibited items are annotated). It mimics a similar testing environment to the real-world scenarios where inspectors often aim at recognizing prohibited items appearing in a very low frequency ({\em e.g.}, $1$ in $1\rm{,} 000$). Unlike GDXray which only contains grayscale images in simple backgrounds, our dataset is much more challenging. Although a color-X-ray scanner assigns various colors to different materials, objects in the containers often suffer a considerable variety in scale, viewpoint, and style, yet a prohibited item may be mixed and overlapped with arbitrary numbers and types of safe items, as shown in Figure~\ref{fig:dataset_examples}.

\begin{figure}[t]
\centering
\includegraphics[width=\linewidth]{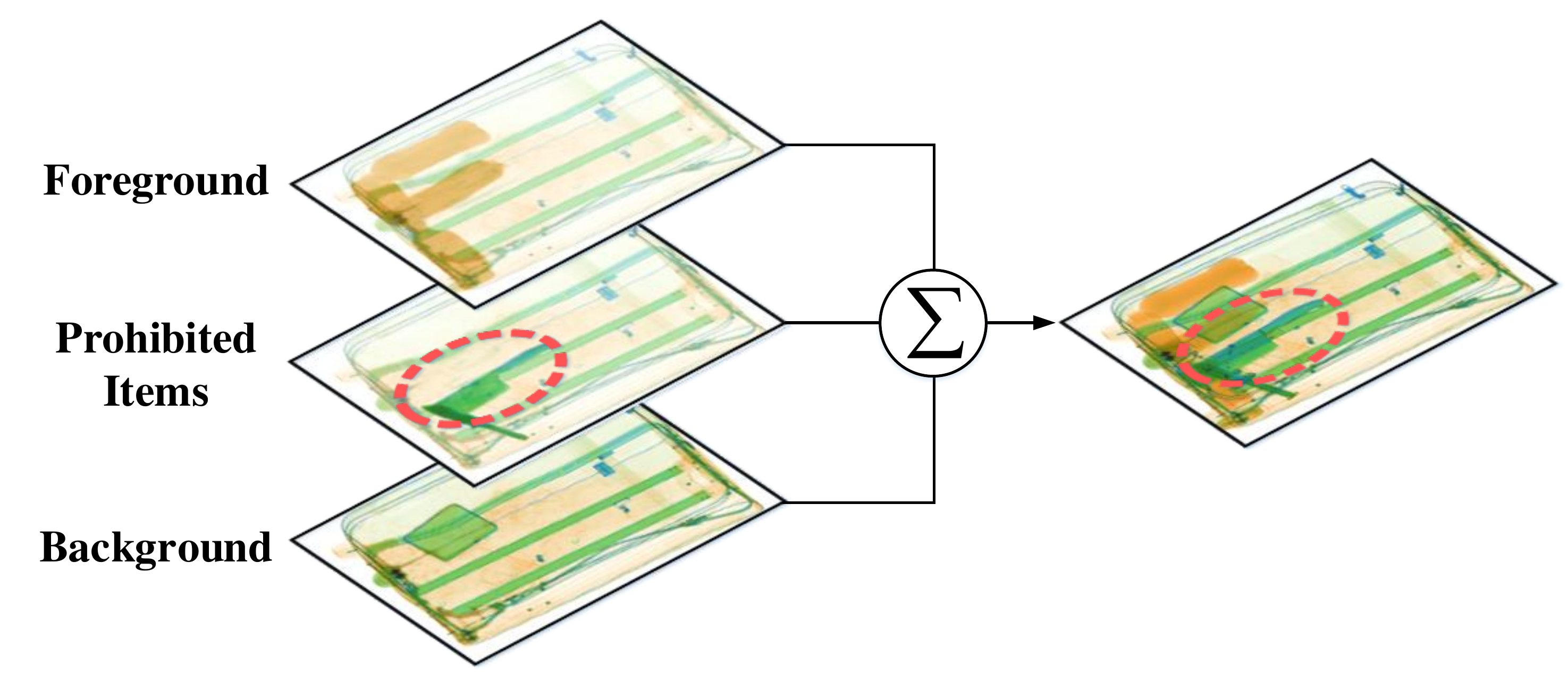}
\caption{An X-ray image is composed of a set of overlapping images, each of which is transparent.  (Best viewed in color).}
\label{fig:overlapping}
\end{figure}

We formulate this problem into an optimization task which, provided a dataset ${\mathcal{D}}={\left\{\left(\mathbf{x}_n,\mathbf{y}_n^*\right)\right\}_{n=1}^N}$, aims at minimizing the expected loss function between ground-truth and prediction $\left|\mathbf{y}_n^*-\mathbf{f}\!\left(\mathbf{x}_n;\boldsymbol{\theta}\right)\right|^2$. Here $\mathbf{x}_n$ denotes image data and $\mathbf{y}_n^*$ is a $C$-dimensional vector with each index indicating whether a specific class is present in $\mathbf{x}_n$. Based on this framework, we point out a clear difference between natural images and X-ray images. A natural image $\mathbf{x}_n$ often contains only one class $c_n$ and thus can be sampled from a distribution $\mathcal{P}\!\left(\mathbf{x}\mid c_n\right)$. However, an X-ray image is often composed of a set of overlapping images which, provided a multi-class label $\mathbf{y}_n^*$ ($C$ dimensions), can be formulated using a mixture distribution ${\mathbf{x}_n}={{\sum_c}y_{n,c}^*\cdot\mathbf{x}_{n,c}}$ where $\mathbf{x}_{n,c}$ is sampled from a hidden distribution $\mathcal{P}\!\left(\mathbf{x}\mid c\right)$, as shown in Figure~\ref{fig:overlapping}.

We present an approach in the context of deep neural networks to deal with this complex scenario. The key idea is to combine two sources of information, namely, using mid-level features $\mathbf{x}_n$ (most often sampled from a mixture distribution) to determine high-level semantics $\mathbf{y}_n$, and reversely filtering irrelevant information out of $\mathbf{x}_n$ by referring to the information contained in $\mathbf{y}_n$. To this end, we formulate the high-level supervision signals into reversed network connections. To alleviate data imbalance, we introduce a loss-balancing term based on this hierarchy. This leads to the complete pipeline named class-balanced hierarchical refinement (CHR). With $\mathbf{y}_n$ being unobserved, an iterative process is required in optimization, which is computationally expensive in practice. To accelerate, we switch off iteration so that more training data are processed in a unit time period. In testing, CHR fuses visual information from different stages towards higher recognition accuracy, yet remains efficient in computation.

We evaluate CHR on the SIXray with different ratios of positive/negative samples. CHR reports significantly higher classification performance over various baselines, {\em i.e.}, different network backbones, demonstrating the effectiveness of using high-level cues to assist mid-level features. In addition, we verify the necessity of adding the class-balanced loss term as we observe more significant improvement on less balanced training data. Last but not least, we provide annotations of prohibited items at the bounding box level in the testing set, and apply the class activation mapping (CAM) algorithm~\cite{zhou2016learning} as a baseline for weakly-supervised object localization.

The major contributions of this work are two-fold. ($1$) We provide a benchmark for future research in this challenging vision task. ($2$) We present an approach named CHR, which integrates multi-level visual cues and achieves class balance in the hierarchical structure.

\section{Related Work}

\label{sec:related_work}

\subsection{X-ray Images and Benchmarks}
\label{sec:related_work:benchmarks}

X-ray images are captured by irradiating the objects with X-ray and rendering them with pseudo colors according to their spectral absorption rates. Therefore, in X-ray images, objects made of the same material are assigned with very similar colors, {\em e.g.}, metals are often shown in blue while impenetrable objects are often shown in red. Besides, the most significant difference between X-ray and natural images lies in object overlapping, because X-ray is often applied in the scenarios that some objects may heavily occlude others, {\em e.g.}, in a {\em baggage}, personal items are often stacked randomly. This property brings a new challenge to computer vision algorithms, while the traditional difficulties persist, {\em e.g.}, scale and viewpoint variance, intra-class variance and inter-class similarity, {\em etc.}, as widely observed in other object localization benchmarks like PascalVOC~\cite{2010pascal} and MS-COCO~\cite{2014microsoft}.

Researchers designed much work to deal with these difficulties and also approach the promising commercial value after them~\cite{a3}\cite{franzel2012object}\cite{mery2015object}\cite{roomi2012detection}\cite{turcsany2013improving}. But unfortunately, very few X-ray datasets have been published for research purposes. A recently released benchmark, GDXray~\cite{mery2015gdxray}, contains three major categories of prohibited items including {\em gun}, {\em shuriken} and {\em razor blade}. However, images in GDXray were provided with few background clutters as well as overlaps, thus, it becomes considerably easy to recognizing these images and/or detecting the objects within. In addition, the relatively small number of negative samples (images not containing prohibited items) ease the algorithm in both training and testing stages. ChestXray8~\cite{wang2017chestx} is a large-scale chest X-ray corpus for medical imaging analysis. Different from our scenario, objects in these images are rarely overlapping with each other.

\subsection{Object Recognition and Localization}
\label{sec:related_work:algorithms}

The research field of object recognition has been dominated by deep learning approaches. With the availability of large-scale datasets~\cite{2012imagenet} and powerful computational resources, researchers are able to design and optimize very deep neural networks~\cite{2012imagenet}\cite{szegedy2017inception}\cite{Simonyan2014Very}\cite{Szegedy2015Going}\cite{he2016deep}\cite{huang2017densely} to learn visual patterns in a hierarchical manner. In the scenario that each image may contain more than one objects, there are typically two types of localization methods. The first one worked on the image level which produces a score for each class indicating its presence or absence~\cite{zhou2016learning}. The second one instead worked on the object level, and produced a bounding box as well as a class label for each object individually~\cite{girshick2014rich}\cite{girshick2015fast}\cite{ren2015faster}\cite{liu2016ssd}\cite{redmon2016you}. The former type often encounters the issues of multi-object classification and training data imbalance~\cite{wang2017chestx}, for which  binary cross entropy (BCE) loss~\cite{BCE2017} as well as class-balancing techniques~\cite{wang2017chestx}\cite{japkowicz2002class} were explored. The second type, on the other hand, was often based on a pipeline that first extracts a number of proposals in the image~\cite{girshick2014rich}\cite{girshick2015fast}\cite{ren2015faster}, and then determines the class of each proposal.

This paper studies image-level recognition, as per-object annotation is missing for training data, while our approach has the ability of object-level localization. This is related to the research in weakly-supervised object localization~\cite{bilen2016weakly}\cite{diba2017weakly}\cite{Tang2017OICR}, or a series of work in localizing objects using

top-down class activation \cite{durand2016weldon}\cite{durand2017wildcat}\cite{zhu2017soft}. There were also efforts about formulating the object localization in multiple instance learning frameworks where convolutional filters behave as detectors
which activate regions of interest on the feature maps \cite{bilen2016weakly}\cite{Ren2016Weakly}\cite{Tang2017OICR}.

In the context of object recognition in X-ray images, researchers realized that these images often contain fewer texture information, yet shape information stands out to be more discriminative. Therefore, in the era of bag-of-visual-word models~\cite{turcsany2013improving}\cite{bacstan2011visual}, the topic of designing effective and efficient handcrafted features is explored in depth~\cite{roomi2012detection}\cite{mery2015object}. As deep learning becomes a standard tool of optimizing complex functions, researchers started to apply it to either extracting compact visual features for X-ray image representation~\cite{a3} or fine-tuning a pre-trained model on X-ray images so that knowledge learned from natural images can be borrowed. This paper mainly focuses on the second approach.

\section{The SIXray Benchmark}
\label{sec:benchmark}

\subsection{Data Acquisition}

We collected a dataset named Security Inspection X-ray (SIXray), which contains a total of $1\rm{,}059\rm{,}231$ X-ray images, and is more than $100$ times larger than the only existing public dataset for the same purpose, {\em i.e.}, the {\em baggage} group of the GDXray dataset~\cite{mery2015gdxray}. These images were collected from several subway stations with the original meta-data indicating the presence or absence of prohibited items. There are six common categories of prohibited items, namely, {\em gun}, {\em knife}, {\em wrench}, {\em pliers}, {\em scissors}, and {\em hammer}. The {\em hammer} class with merely $60$ samples is not used in our experiments.

The distribution of these objects aligns with the real-world scenario, in which there are much fewer positive samples compared to negative samples. A statistics on this dataset is shown in Table~\ref{table_SIXray}. Each image was scaned by security inspection machine , which assigned different colors to objects made of different materials. All images were stored in JPEG format with an average size of $100\mathrm{K}$ pixels.

To study the impact brought by training data imbalance, we construct three subsets of this dataset, and name them as SIXray10, SIXray100 and SIXray1000, respectively, with the number indicating the ratio of negative samples over positive samples. In SIXray10 and SIXray100, all $8\rm{,}929$ positive images are included, and there are exactly $10\times$ and $100\times$ negative images. SIXRay100 has a very close distribution to the real world scenario. To maximally explore the ability of our algorithm to deal with data imbalance, we construct the SIXray1000 dataset by randomly choosing only $1\rm{,}000$ positive images but mixing them with all the $1\rm{,}050\rm{,}302$ negative images. Each subset is further partitioned into a training set and a testing set, with the former containing $80\%$ of the images and the latter containing $20\%$ (the ratio training/testing images is $4:1$).

On the entire dataset, we use the image-level annotations provided by human security inspectors, {\em i.e.}, whether each type of prohibited items is present. In addition, on the {\em testing} sets, we manually add a bounding-box for each prohibited item to evaluate the performance of object localization.

\begin{table}[t]
\centering
\small

\setlength{\tabcolsep}{0.08cm}
\begin{tabular}{|c|c|c|c|c|c|c|}
\cline{1-7}
\multicolumn{7}{|c|}{The SIXray Dataset ($1\rm{,}059\rm{,}231$)} \\
\cline{1-7}
\multicolumn{6}{|c|}{Positive ($8\rm{,}929$)} & \multirow{2}{*}{Negative} \\
\cline{1-6}
Gun & Knife & Wrench & Pliers & Scissors & Hammer & {} \\
\cline{1-7}
$3\rm{,}131$ & $1\rm{,}943$  & $2\rm{,}199$ & $3\rm{,}961$ & $983$ & $ 60 $ & $1\rm{,}050\rm{,}302$ \\
\cline{1-7}
\end{tabular}
\vspace{0.1cm}
\caption{The class distribution of the SIXray dataset. There is another {\em hammer} class with $60$ items, but it is not used due to the small number of samples.}
\label{table_SIXray}
\end{table}

\subsection{Dataset Properties}

The SIXray dataset has several properties which bring difficulties to visual recognition. {\bf First}, these images were mostly obtained from X-ray scans on personal luggage, {\em e.g.}, {\em bags} or {\em suitcases}, in which objects are often randomly stacked. When these items passed an X-ray scan, the penetration property makes it possible to see even the occluded items in the image. This leads to the most important property of this dataset, which we call it {\em overlapping}. Note that GDXray~\cite{mery2015gdxray} does not have such a challenge as there is often only one item in each image. {\bf Second}, prohibited items can appear in many different scales, viewpoints, styles and even subtypes, all of which cause considerable intra-class variation and increase the difficulty of recognition. {\bf Third}, the images can be heavily cluttered yet it is almost impossible to assign all objects especially those non-prohibited ones with a clear class label. Thus, there is noise coming from an open set of objects, which makes it difficult to expect what appears in the background regions. {\bf Fourth and last}, as mentioned above, the positive images (with at least one prohibited item) only occupy a small fraction of this dataset. Without a special treatment, it is easy for the training stage to bias towards the negative class, as simply guessing a negative label yields sufficiently high accuracy. This raises a challenge to training stability.

In the following section, we present our approach which takes these properties into consideration, especially the first and fourth properties which are specific to this dataset.

\section{Our Approach}
\label{sec:approach}

\subsection{Motivation and Formulation}
\label{sec:approach:motivation}

As observed in the previous section, a significant characteristic of X-ray images lies in that objects are overlapped with each other. Note that overlapping is different from occlusion in which the rear object is invisible. Instead, as X-ray is penetrable, both front and rear objects are visible in the image. This is named the {\em penetration assumption}, based on which we use a mixture model to formulate these data.

Let there be $C$ classes of possible items appearing in the dataset, with an index set of $\left\{1,2,\ldots,C\right\}$. Among them, $C'$ classes are considered prohibited, {\em e.g.}, in the SIXray dataset, ${C'}={5}$. Without loss of generality, we assign them with the class index of $1,2,\ldots,C'$. Let the dataset $\mathcal{D}$ contain $N$ images. For each input image $\mathbf{x}_n$, our goal is to obtain a $C$-dimensional vector $\mathbf{y}_n$ for each $\mathbf{x}_n$, each dimension in which, $y_{n,c}$, is either $0$ or $1$, with $1$ indicating the specified prohibited item is present in this image and $0$ vice versa. Note that the ground-truth of $\mathbf{y}_n^\star$ only exists for the first $C'$ dimensions, while others remain unobserved.

To obtain a mathematical formulation of $\mathbf{x}_n$, we assume that it is composed of $C$ sub-images $\mathbf{x}_{n,c}$, each of which corresponds to a specified class $c$ and is sampled from a conditional distribution ${\mathcal{P}_c}\doteq{\mathcal{P}\!\left(\mathbf{x}\mid c\right)}$. Then, based on the {\em penetration assumption}, each image can be written as:
\begin{equation}
\label{eqn:image_composition}
{\mathbf{x}_n}\approx{{\sum_{c=1}^C}y_{n,c}\cdot\mathbf{x}_{n,c},\quad\mathbf{x}_{n,c}\sim\mathcal{P}_{c}}.
\end{equation}
This formulation is of course not accurate as we ignore the overlapping relationship between objects as well as the order that objects are stacked, but it serves as an approximate formulation of how overlapping impacts image data.

Our goal is to learn a discriminative function ${\mathbf{y}_n}={\mathbf{f}\!\left(\mathbf{x}_n;\boldsymbol{\theta}\right)}$ to predict the image label. Since the object of interest may appear in various scales. In order to recognize and further detect it, a popular choice~\cite{ke2017srn}\cite{lin2017feature} is to combine multi-stage visual information. Here we simply consider feature vectors extracted from $L$ different layers, the $l$-th of which is denoted as $\mathbf{x}_n^{\left(l\right)}$. A regular solution is to train a classifier beyond each layer, ${\mathbf{y}_n^{\left(l\right)}}={\mathbf{h}^{\left(l\right)}\!\left(\mathbf{x}_n^{\left(l\right)};\boldsymbol{\xi}^{\left(l\right)}\right)}$, using the ground-truth signal $\mathbf{y}_n^\star$ as supervision. In the testing stage, we fuse all $\mathbf{y}_n^{\left(l\right)}$ as the final output, {\em i.e.}, ${\mathbf{y}_n}={{\sum_{l=1}^L}\mathbf{y}_n^{\left(l\right)}}$.

However, we note a significant weakness of this model, which comes from the penetration assumption, {\em i.e.}, Eqn~\eqref{eqn:image_composition}, applied to mid-level features\footnote{Eqn~\eqref{eqn:image_composition} fits mid-level features best, because low-level features ({\em e.g.}, raw image pixels) are often largely impacted by small noise, in both case it is learning the class-conditional distribution ${\mathcal{P}_c}\doteq{\mathcal{P}\!\left(\mathbf{x}\mid c\right)}$ suffers a higher difficulty. Similarly, the very last layers ({\em e.g.}, containing class-specific {\em logits}) are less likely to be additive as in Eqn~\eqref{eqn:image_composition}.}. This is to say, each $\mathbf{x}_n^{\left(l\right)}$ is the composition of sub-images sampled from different classes, including those items of no interest, and thus $\mathbf{h}^{\left(l\right)}\!\left(\mathbf{x}_n^{\left(l\right)};\boldsymbol{\xi}^{\left(l\right)}\right)$ may be distracted. A reasonable idea is to refine $\mathbf{x}_n^{\left(l\right)}$ to get rid of these irrelevant information. This is achieved by a function $\mathbf{g}^{\left(l\right)}\!\left(\mathbf{x}_n^{\left(l\right)},\mathbf{y}_n;\boldsymbol{\tau}^{\left(l\right)}\right)$, which shares the same dimensionality with $\mathbf{x}_n^{\left(l\right)}$. Summarizing these contents yields the following optimization problem:
\begin{align}
\label{eqn:optimization}
{\boldsymbol{\theta}^\star,\boldsymbol{\xi}^\star,\boldsymbol{\tau}^\star} & ={\arg\min_{\boldsymbol{\theta},\boldsymbol{\xi},\boldsymbol{\tau}}\mathbb{E}_{\mathbf{x}_n\in\mathcal{D}}{\sum_{l=1}^L}\mathcal{L}_n^{\left(l\right)}},\quad\mathrm{where}\\
\label{eqn:classification_loss}
{\mathcal{L}_n^{\left(l\right)}} & ={\mathcal{L}\!\left\{\mathbf{y}_n^\star,\mathbf{h}^{\left(l\right)}\!\left(\tilde{\mathbf{x}}_n^{\left(l\right)};\boldsymbol{\xi}^{\left(l\right)}\right)\right\}},\\
\label{eqn:refinement}
{\tilde{\mathbf{x}}_n^{\left(l\right)}} & ={\mathbf{g}^{\left(l\right)}\!\left(\mathbf{x}_n^{\left(l\right)},\mathbf{y}_n;\boldsymbol{\tau}^{\left(l\right)}\right)},\quad\mathrm{and}\\
\label{eqn:overall_output}
{\mathbf{y}_n} & ={\frac{1}{L}\cdot{\sum_{l=1}^L}\mathbf{h}^{\left(l\right)}\!\left(\tilde{\mathbf{x}}_n^{\left(l\right)};\boldsymbol{\xi}^{\left(l\right)}\right)}.
\end{align}
Here $\mathcal{L}\!\left\{\cdot,\cdot\right\}$ is a loss function which is discussed in details later. The above formulae define a recurrent model, in which $\mathbf{y}_n$ cannot be observed even in the training stage. The standard way of optimization involves iteration, in which we start with an $\mathbf{x}_n$ sampled from $\mathcal{D}$ and any $\mathbf{y}_n$ (in the training process, the first $C'$ dimensions are provided by ground-truth and other $C-C'$ dimensions can be randomly initialized). We first compute $\mathbf{x}_n^{\left(l\right)}$ for each $l$ accordingly, and use it to compute the first version of ${\mathbf{y}_n^{\left(l\right)}}={\mathbf{h}^{\left(l\right)}\!\left(\mathbf{x}_n^{\left(l\right)};\boldsymbol{\xi}^{\left(l\right)}\right)}$. In each round, we compute $\mathbf{y}_n$ and use it to compute $\mathbf{g}^{\left(l\right)}\!\left(\mathbf{x}_n^{\left(l\right)},\mathbf{y}_n;\boldsymbol{\tau}^{\left(l\right)}\right)$ so that $\mathbf{x}_n^{\left(l\right)}$ is updated as $\tilde{\mathbf{x}}_n^{\left(l\right)}$. Within this process, parameters $\boldsymbol{\xi}^{\left(l\right)}$ and $\boldsymbol{\tau}^{\left(l\right)}$ are updated accordingly with ground-truth $\mathbf{y}_n^\star$ and gradient back-propagation. This iteration continues until convergence or a maximal number of rounds is achieved\footnote{Here are some side notes. It has been widely believed that a deep network is able to fit training data sampled from one-class distributions, {\em e.g.}, each sample contains only one object in class $c_n$, so that $\mathbf{x}_n$ is sampled from $\mathcal{P}_{c_n}$. In such scenarios, $\mathbf{y}_n$ as a one-hot vector is relatively easy to estimate and thus iteration is not required. This is the reason that deep networks produced satisfying performance in the GDXray dataset~\cite{mery2015gdxray} in which most images contain only one object.}.

\begin{figure*}[t]
\centering
\includegraphics[width=1.0\linewidth]{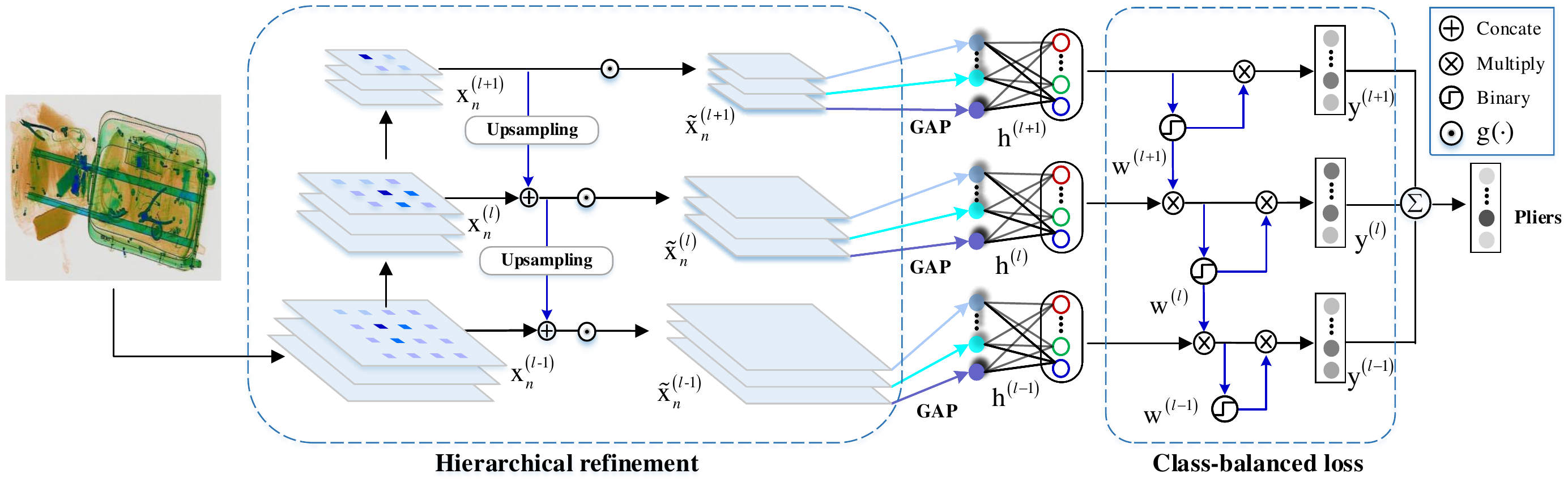}
\caption{The overall architecture of the proposed class-balanced hierarchical refinement (CHR) approach (best viewed in color). The network backbone $\mathbf{f}\!\left(\mathbf{x}_n;\boldsymbol{\theta}\right)$ is shown on the leftmost column, from which $L$ layers are chosen as feature extractors. For simplicity, we show an example with ${L}={3}$. Each $\tilde{\mathbf{x}}_n^{\left(l\right)}$, ${l}>{1}$, is up-sampled and concatenated with $\mathbf{x}_n^{\left(l-1\right)}$ and fed into a refinement function that simulates ${\tilde{\mathbf{x}}_n^{\left(l-1\right)}}={\mathbf{g}\!\left(\mathbf{x}_n^{\left(l-1\right)},\mathbf{x}_n^{\left(l\right)};\boldsymbol{\tau}^{\left(l-1\right)}\right)}$, and $\tilde{\mathbf{x}}_n^{\left(l-1\right)}$ is sent into $\mathbf{h}^{\left(l-1\right)}\!\left(\tilde{\mathbf{x}}_n^{\left(l-1\right)};\boldsymbol{\xi}^{\left(l-1\right)}\right)$ for classification. GAP denotes global average pooling. A class-balancing loss is built upon the same hierarchy, on which mid-level negative samples are filtered out using high-level cues. }
\label{fig:flowchart}
\end{figure*}

\subsection{Approximation with Hierarchical Refinement}
\label{sec:approach:optimization}

In practice, however, the above formulation has two major drawbacks. The first one lies in the inaccuracy of generative models. We expect a model $\mathbf{g}^{\left(l\right)}\!\left(\cdot\right)$ to eliminate the components in $\mathbf{x}_n^{\left(l\right)}$ that correspond to the non-targeted classes in $\mathbf{y}_n$. This is increasingly difficult especially when the $\mathbf{x}_n^{\left(l\right)}$ is far from $\mathbf{y}_n$. So, we assume that $\mathbf{x}_n^{\left(l\right)}$ only receives supervision signals from $\mathbf{x}_n^{\left(l+1\right)}$, which is much closer than $\mathbf{y}$, while $\mathbf{x}_n^{\left(l+1\right)}$ continues to receive information from $\mathbf{x}_n^{\left(l+2\right)}$ and this process continues until $\mathbf{y}_n$ is reached. In implementation, this implies that reversed connections only emerge between neighboring feature layers. Here an exception happens at the last feature layer, $\mathbf{x}_n^{\left(L\right)}$, which is connected to $\mathbf{y}_n$ via a classifier $\mathbf{h}^{\left(L\right)}\!\left(\cdot\right)$. Since direct supervisions have already been provided by this classifier, we ignore the connection between $\mathbf{y}_n$ and $\mathbf{x}_n^{\left(L\right)}$, leaving a total of $L-1$ connections between $\mathbf{x}_n^{\left(l\right)}$ and $\mathbf{x}_n^{\left(l+1\right)}$, for ${l}={1,2,\ldots,L-1}$. This is to say, $\mathbf{g}\!\left(\mathbf{x}_n^{\left(l\right)},\mathbf{y}_n;\boldsymbol{\xi}^{\left(l\right)}\right)$ is replaced by $\mathbf{g}\!\left(\mathbf{x}_n^{\left(l\right)},\mathbf{x}_n^{\left(l+1\right)};\boldsymbol{\xi}^{\left(l\right)}\right)$. Nevertheless, $\mathbf{x}_n^{\left(l\right)}$ can still obtain supervision signals from $\mathbf{y}_n$ in an indirect manner, {\em i.e.}, via a few intermediate steps. This is named the hierarchical refinement strategy.

Implementation details are illustrated in Figure~\ref{fig:flowchart}. We start with ${\tilde{\mathbf{x}}_n^{\left(L\right)}}\equiv{\mathbf{x}_n^{\left(L\right)}}$, the feature extracted from the top layer. It is concatenated with the feature at the previous stage, $\mathbf{x}_n^{\left(L-1\right)}$, before which it is up-sampled if necessary. The concatenated feature is then fed into $\mathbf{g}^{\left(L-1\right)}\!\left(\mathbf{x}_n^{\left(L-1\right)},\mathbf{x}_n^{\left(L\right)};\boldsymbol{\tau}^{\left(L-1\right)}\right)$ to produce $\tilde{\mathbf{x}}_n^{\left(L-1\right)}$. This process continues until $\tilde{\mathbf{x}}_n^{\left(1\right)}$ is obtained. Each $\tilde{\mathbf{x}}_n^{\left(l\right)}$, ${l}={1,2,\ldots,L}$, is sent into the corresponding classifier $\mathbf{h}^{\left(l\right)}\!\left(\tilde{\mathbf{x}}_n^{\left(l\right)};\boldsymbol{\xi}^{\left(l\right)}\right)$ to obtain $\mathbf{y}_n^{\left(l\right)}$. All $\mathbf{y}_n^{\left(l\right)}$ are averaged into the final output and supervised by $\mathbf{y}^\star$.

The second drawback is the slowness of an iterative optimization. To accelerate, we switch off iteration so that each case ${\mathbf{x}_n}\in{\mathcal{D}}$ is forward-propagated and back-propagated only once, and the updated parameters $\boldsymbol{\theta}$, $\boldsymbol{\xi}^{\left(l\right)}$ and $\boldsymbol{\tau}^{\left(l\right)}$ are directly applied to another case sampled from $\mathcal{D}$. This can be understood as stochastic gradient descent on $\mathcal{D}$. In practice, this allows us to sample more data in the same period of time, and thus improve training efficiency.

\subsection{Class-Balanced Loss}
\label{sec:approach:loss}

Here we study the impact of the loss function, {\em i.e.}, Eqn~\eqref{eqn:classification_loss}, in the training process. In this specific problem, {\em i.e.}, prohibited item discovery, there are much fewer positive training samples (at least one prohibited item is labeled) than negative ones. This makes regular loss functions such as the Euclidean loss ${\mathcal{L}\!\left\{\mathbf{y}_n^\star,\mathbf{y}_n\right\}}={\left|\mathbf{y}_n^\star-\mathbf{y}_n\right|^2}$ and the Binary Cross-Entropy (BCE) loss ${\mathcal{L}\!\left\{\mathbf{y}_n^\star,\mathbf{y}_n\right\}}={-\left[\mathbf{y}_n^{\star\top}\log\mathbf{y}_n+\left(1-\mathbf{y}_n^\star\right)^\top\log\!\left(1-\mathbf{y}_n\right)\right]}$ less effective, because the network can heavily bias towards negative examples (because simply guessing all training samples to be negative leads to a very low loss function) and, consequently, the recall becomes considerably low. A reasonable solution is to slightly change the loss function so as to equivalently reduce the number of negative training data~\cite{wang2017chestx}. Here we combine this approach in the context of hierarchical refinement which once again takes advantage of high-level supervision to guide mid-level features.

The proposed loss function works in a mini-batch ${\mathcal{B}}\subset{\mathcal{D}}$. For each case $\mathbf{x}_n$ with $\mathbf{y}_n$, we have a few stages defined previously, each of which produces a feature $\mathbf{x}_n^{\left(l\right)}$ followed by a prediction $\mathbf{y}_n^{\left(l\right)}$. We add a binary weight vector, denoted by $\mathbf{w}_n^{\left(l\right)}$, measuring whether each class in $\mathbf{y}_n^{\left(l\right)}$ contributes to the loss function. Thus, Eqn~\eqref{eqn:classification_loss} becomes:
\begin{equation}
\label{eqn:weighted_loss}
{\mathcal{L}_n^{\left(l\right)}}={\mathbf{w}_n^{\left(l\right)\top}\cdot\mathbf{E}\!\left(\mathbf{y}_n^\star,\mathbf{y}_n^{\left(l\right)}\right)},
\end{equation}
where $\mathbf{E}\!\left(\mathbf{y}_n^\star,\mathbf{y}_n^{\left(l\right)}\right)$ is the loss vector, ${\mathbf{E}\!\left(\mathbf{y}_n^\star,\mathbf{y}_n^{\left(l\right)}\right)}={-\left[\mathbf{y}_n^{\star}\odot\log\mathbf{y}_n^{\left(l\right)}+\left(1-\mathbf{y}_n^\star\right)\odot\log\!\left(1-\mathbf{y}_n^{\left(l\right)}\right)\right]}$, and $\odot$ denotes element-wise multiplication.

It remains to define $\mathbf{w}_n^{\left(l\right)}$ for each $\mathbf{y}_n^{\left(l\right)}$. In the highest ($L$-th) level, $\mathbf{w}_n^{\left(L\right)}$ directly measures whether each class, or each dimension in $\mathbf{y}_n^{\left(l\right)}$, has to be considered. This conditional variable is always true for each class with a positive label, while for that with a negative label, it is true only if the prediction is larger than a fixed threshold $\varepsilon$. In each of the lower levels, a class is considered if the above judgment returns true, as well as all the higher levels support this -- in other words, if a class is switched off at some level, it will never be considered in each of the lower levels. This is based on the assumption that high-level features are more reliable in determining which classes are present and which are absent, while low-level features may produce false positives due to various reasons.

\begin{figure}[t]
\centering
\includegraphics[width=\linewidth]{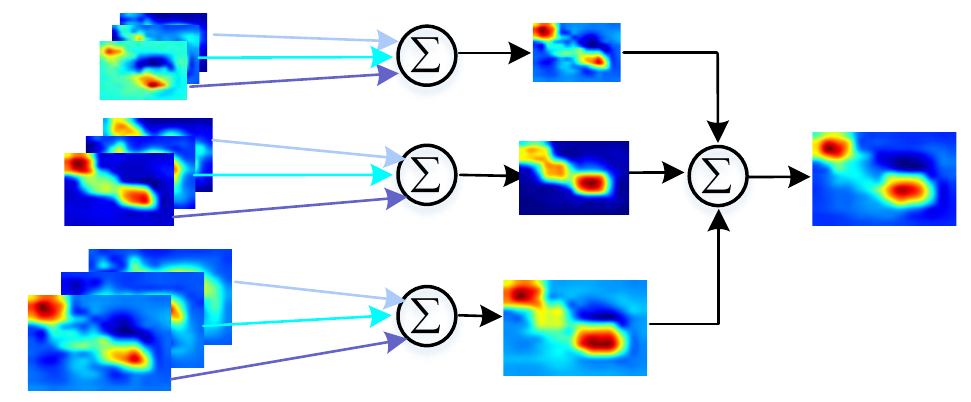}
\caption{Discriminative prohibited item localization with hierarchical features.  (Best viewed in color).}
\label{fig:formulation}
\vspace{-0.2cm}
\end{figure}

Replacing Eqn~\eqref{eqn:classification_loss} with Eqn~\eqref{eqn:weighted_loss} gives the complete class-balanced hierarchical refinement (CHR) approach. In the training process, each $\mathcal{L}_n^{\left(l\right)}$ is computed individually and averaged for gradient back-propagation. In the testing stage, we directly average all $\mathbf{y}_n^{\left(l\right)}$ for the final prediction. Please refer to Figure~\ref{fig:formulation} for details.

\section{Experiments}
\label{sec:experiments}

\subsection{Setting and Baselines}
\label{sec:experiments:setting_baselines}

We use all three subsets, namely, SIXRay10, SIXRay100 and SIXRay1000, to evaluate different approaches. In each subset, all models are optimized on $80\%$ training data, and evaluated on the remaining $20\%$ testing data. These data splits are random but consistent for all competitors.

We evaluate both image-level classification mean Average Precision and object-level localization accuracies, for the second goal we manually labeled all prohibited items with a bounding-box in the testing images. For image classification, we apply the evaluation metric in the PascalVOC image classification task~\cite{2010pascal}, which works on each class individually -- all testing images are ranked by the confidence of containing the specified object, and the mean average precision (mAP) is computed. For object localization, we follow~\cite{zhang2018top} to compute the accuracy of pointing localization. A hit is counted if the pixel of the maximum response falls within one of the ground-truth bounding-boxes of the specified object, otherwise a missed is counted. Thus, each class has a localization accuracy computed by $\frac{\#\mathrm{Hits}}{\#\mathrm{Hits}+\#\mathrm{Misses}}$. For both tasks, we also report the overall performance which is the average over all five classes.

We investigate five popular backbones, including ResNets~\cite{he2016deep} with $34$, $50$ and $101$ layers, Inception-v3~\cite{szegedy2016rethinking}, and densenet with 121 layers. We follow the conventions to setup all these networks, and CHR is applied to each of them using ${L}={3}$ -- three pooling layers with different spatial resolutions ({\em e.g.}, in ResNets, $28\times28$, $14\times14$, and $7\times7$) are used as features. It is of course possible to increase $L$ by adding more features, yet in practice we find ${L}={3}$ is sufficient to provide complementary information.

\begin{table*}[!t]
\footnotesize
\begin{center}
\setlength{\tabcolsep}{0.08cm}
\begin{tabular}{|l|c|c|c|c|c|c|c|c|c|c|c|c|c|c|c|c|c|c|}
\hline
Method & \multicolumn{3}{c|}{Gun} & \multicolumn{3}{c|}{Knife} & \multicolumn{3}{c|}{Wrench} & \multicolumn{3}{c|}{Pliers} & \multicolumn{3}{c|}{Scissors} & \multicolumn{3}{c|}{mean}\\
\hline\hline
ResNet34\cite{he2016deep}
& $89.71$ & $83.06$ & $72.05$   & $85.46$ & $78.75$ & $56.42$ & $62.48$ & $30.49$ & $16.47$ & $83.50$ & $55.24$ & $14.24$  & $52.99$ & $16.14$ & $7.12$ & $74.83$ & $52.74$ & $33.26$  \\
\hline
ResNet34+CHR
& $87.16$ & $81.96$ & $73.35$ & $87.17$ & $77.70$ & $60.46$ & $64.31$ & $36.85$ & $23.72$ & $85.79$ & $64.56$ & $17.98$ & $61.58$ & $14.49$ & $18.19$ & $77.20$ & $55.11$ & $38.74$ \\
\hline
ResNet50\cite{he2016deep}
& $90.64$ & $84.75$ & $74.19$ & $87.82$ & $77.92$ & $59.82$ & $63.62$ & $28.49$ & $16.03$ & $84.80$ & $50.53$ & $16.59$ & $57.35$ & $19.39$ & $2.87$ & $76.85$ & $52.22$ & $33.90$ \\
\hline
ResNet50+CHR
& $87.55$ & $82.64$ & $73.43$ & $86.38$ & $79.60$ & $61.32$ & $69.12$ & $41.19$ & $18.88$ & $85.72$ & $58.02$ & $12.32$ & $60.91$ & $27.89$ & $19.03$ & $77.94$ & $57.87$ & $37.00$ \\
\hline
ResNet101\cite{he2016deep}
& $87.65$ & $82.83$ & $76.04$ & $84.26$ & $76.16$ & $63.53$  & $69.33$ & $35.59$ & $13.65$ & $85.29$ & $54.82$ & $15.57$ & $60.39$ & $20.63$ & $11.28$ & $77.38$ & $54.01$ & $36.01$  \\
\hline
ResNet101+CHR
& $85.45$ & $83.25$ & $75.38$ & $87.21$ & $77.53$ & $64.80$ & $71.23$ & $42.02$ & $15.27$ & $88.28$ & $68.01$ & $19.02$ & $64.68$ & $32.33$ & $16.21$ & $79.37$ & $60.63$ & $38.14$  \\
\hline\hline
Inception-v3\cite{szegedy2016rethinking}
& $90.05$ & $81.18$ & $75.52$ & $83.80$ & $77.28$ & $56.33$ & $68.11$ & $32.47$ & $24.01$ & $84.45$ & $66.89$ & $16.75$ & $58.66$ & $22.63$ & $20.72$ & $77.01$ & $56.09$ & $38.67$ \\
\hline
Inception-v3+CHR
& $88.90$ & $79.22$ & $76.91$ & $87.23$ & $73.48$ & $61.29$ & $69.47$ & $37.20$ & $29.60$ & $86.37$ & $69.01$ & $19.11$ & $65.50$ & $31.81$ & $47.56$ & $79.49$ & $58.15$ & $46.89$ \\
\hline
DenseNet\cite{huang2017densely}
& $87.36$ & $83.23$ & $75.00$ & $87.71$ & $77.24$ & $65.55$ & $64.15$ & $37.72$ & $23.57$ & $87.63$ & $62.69$ & $18.09$ & $59.95$ & $24.89$ & $14.18$ & $77.36$ & $57.15$ & $39.28$\\
\hline
DenseNet+CHR
& $87.05$ & $82.06$ & $74.87$ & $85.89$ & $78.75$ & $71.23$ & $70.47$ & $43.22$ & $29.79$ & $88.34$ & $66.75$ & $21.57$ & $66.07$ & $28.80$ & $44.27$ & $79.56$ & $59.92$ & $48.36$ \\
\cline{1-19}
\end{tabular}
\end{center}
\caption{Classification mean Average Precision ($\%$) on subsets of SIXray (each cell, left to right: SIXray10, SIXray100, SIXray1000).}
\vspace{-0.1cm}
\label{tab:classification}
\end{table*}

\begin{table*}[!t]
\footnotesize
\begin{center}
\setlength{\tabcolsep}{0.08cm}
\begin{tabular}{|l|c|c|c|c|c|c|c|c|c|c|c|c|c|c|c|c|c|c|}
\hline
Method & \multicolumn{3}{c|}{Gun} & \multicolumn{3}{c|}{Knife} & \multicolumn{3}{c|}{Wrench} & \multicolumn{3}{c|}{Pliers} & \multicolumn{3}{c|}{Scissors} & \multicolumn{3}{c|}{mean}\\
\hline\hline
ResNet34\cite{he2016deep}
& $71.60$ & $50.62$ & $53.93$ & $51.28$ & $55.38$ & $38.97$ & $43.32$ & $26.74$ & $22.46$ & $68.88$ & $34.54$ & $13.69$ & $22.16$ & $7.95$ & $6.82$ & $51.45$ & $35.05$ & $27.17$  \\
\hline
ResNet34+CHR
& $75.62$ & $60.19$ & $70.41$ & $55.38$ & $63.08$ & $26.15$ & $52.41$ & $35.83$ & $37.97$ & $58.44$ & $53.70$ & $25.10$ & $19.32$ & $0.00$ & $2.27$ & $52.23$ & $42.56$ & $32.38$ \\
\hline
ResNet50\cite{he2016deep}
& $63.89$ & $47.53$ & $42.32$ & $57.44$ & $52.82$ & $48.72$ & $49.73$ & $28.34$ & $19.79$ & $68.88$ & $39.85$ & $19.77$ & $17.05$ & $1.70$ & $2.84$ & $51.40$ & $34.05$ & $26.69$\\
\hline
ResNet50+CHR
& $68.83$ & $57.72$ & $60.67$ & $58.46$ & $49.23$ & $37.44$ & $54.01$ & $41.18$ & $22.46$ & $77.04$ & $49.91$ & $20.91$ & $15.91$ & $15.34$ & $13.64$ & $54.85$ & $42.67$ & $31.02$ \\
\hline
ResNet101\cite{he2016deep}
& $73.77$ & $73.15$ & $70.41$ & $65.13$ & $64.10$ & $60.00$ & $28.34$ & $25.13$ & $15.51$ & $62.24$ & $31.50$ & $14.07$ & $21.02$ & $11.36$ & $5.68$ & $50.10$ & $41.05$ & $33.13$  \\
\hline
ResNet101+CHR
& $80.86$ & $79.32$ & $79.03$ & $73.85$ & $69.23$ & $61.54$ & $52.41$ & $27.81$ & $21.93$ & $9.30$ & $48.39$ & $17.11$ & $40.34$ & $6.25$ & $19.32$ & $51.35$ & $46.20$ & $39.78$  \\
\hline
Inception-v3\cite{szegedy2016rethinking}
& $79.94$ & $64.81$ & $71.16$ & $75.38$ & $65.64$ & $52.31$ & $59.36$ & $40.11$ & $7.49$ & $59.58$ & $32.83$ & $18.63$ & $40.34$ & $26.14$ & $1.70$ & $62.92$ & $45.91$ & $30.26$ \\
\hline
Inception-v3+CHR
& $78.70$ & $67.59$ & $73.41$ & $74.36$ & $63.08$ & $41.54$ & $52.41$ & $23.53$ & $23.53$ & $59.96$ & $54.27$ & $7.60$ & $52.27$ & $39.20$ & $11.36$ & $63.54$ & $49.53$ & $31.49$ \\
\hline
DenseNet\cite{huang2017densely}
& $74.38$ & $71.60$ & $58.05$ & $71.28$ & $62.05$ & $56.92$ & $59.89$ & $24.60$ & $26.20$ & $71.54$ & $55.60$ & $20.53$ & $35.23$ & $9.66$ & $11.36$ & $62.46$ & $44.70$ & $34.61$\\
\hline
DenseNet+CHR
& $79.01$ & $78.40$ & $76.78$ & $76.92$ & $62.56$ & $57.95$ & $59.36$ & $41.71$ & $39.04$ & $72.49$ & $63.76$ & $39.92$ & $40.34$ & $5.11$ & $5.68$ & $65.62$ & $50.31$ & $43.87$ \\
\cline{1-19}
\end{tabular}
\end{center}
\caption{Localization accuracy ($\%$) on subsets of SIXray (each cell, left to right: SIXray10, SIXray100, SIXray1000).}
\vspace{-0.1cm}
\label{tab_localization}
\end{table*}

\subsection{Classification: Quantitative Results}
\label{sec:experiments:classification_results}

We first investigate the overall (averaged over five classes) image classification results which are summarized in Table~\ref{tab:classification}. CHR achieves consistent mean Average Precision  gain beyond all network backbones as well as in all different subsets, {\em i.e.}, SIXray10, SIXray100 and SIXray1000.

We observe that CHR works better in deeper networks, which is also observed in experiments, {\em e.g.}, on top of Inception-v3 and DenseNet, the absolute improvement over SIXRay1000 is $8.22\%$ and $9.08\%$, respectively.

We next observe five types of objects individually. The benefit brought by CHR is different from class to class. Take the DenseNet as an example. When it is aimed at finding {\em gun}, classification performance is not boosted in all subsets, while we observe significant gains over all the other classes, especially for {\em scissors}, the accuracy is improved by an impressive amount of $30\%$. We can see in Table {\color{red}1} that the training samples of {\em scissors} is the least among all five prohibited items, for which reason the baseline suffers significant bias in the training stage. CHR, by introducing hierarchical signals for supervision, largely alleviates this bias.

\begin{figure}[t]
\centering
\includegraphics[width=1\linewidth]{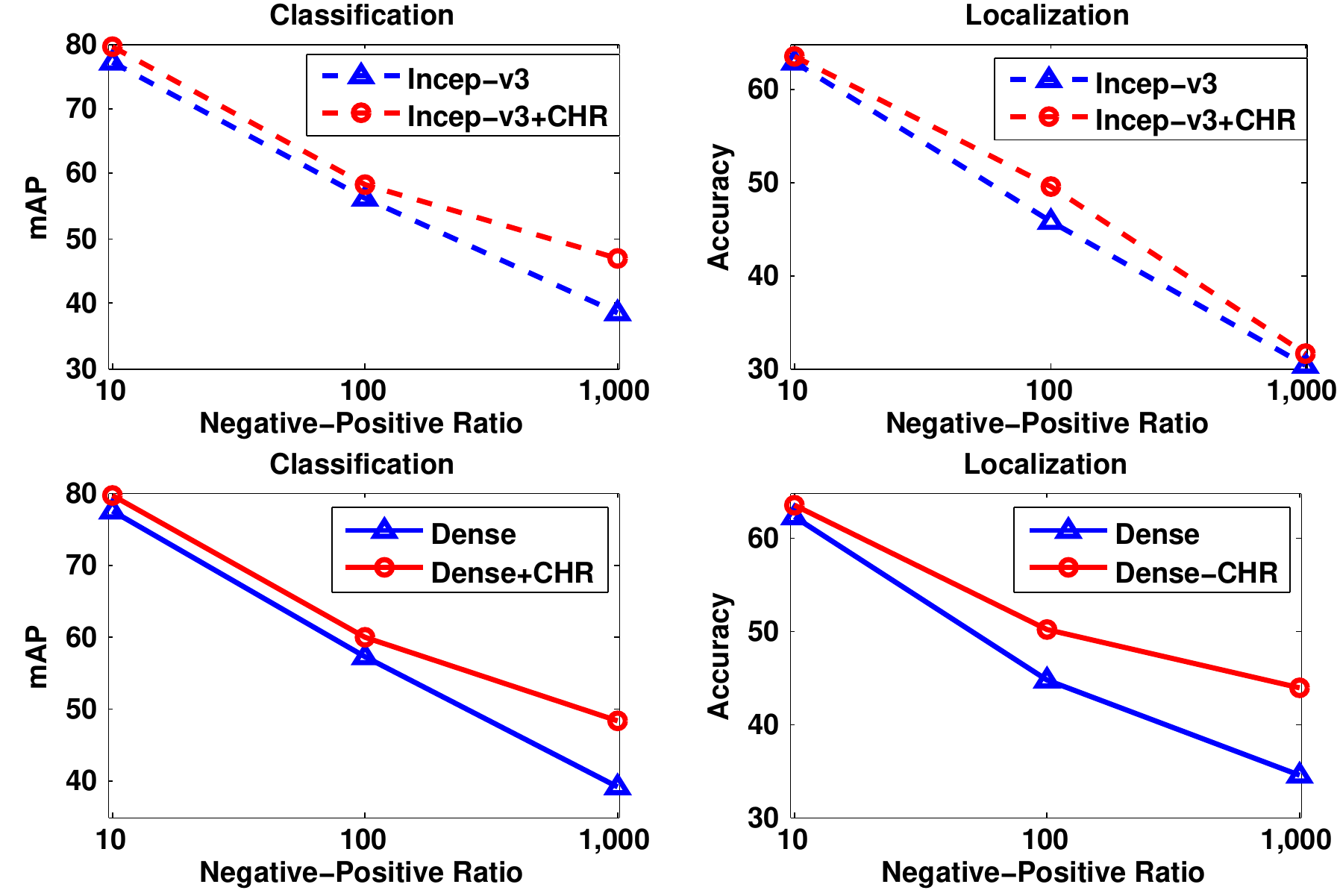}
\caption{The overall accuracy gain of CHR becomes more significant in the subsets with larger negative-positive ratios.}
\vspace{-0.5cm}
\label{fig:imbalance_analysis}
\end{figure}

Finally, we study the issue of data imbalance over different subsets. Recall that the ratio of negative over positive images is $10$, $100$ and $1000$, respectively. From Figure~\ref{fig:imbalance_analysis}, we can see that the performance gain goes up with data imbalance, which, as analyzed in Section~\ref{sec:experiments:ablation}, comes from our special treatment towards class balancing.

\subsection{Localization: Quantitative Results}
\label{sec:experiments:localization_results}

To verify that CHR is not over-tuned to image classification, we attach the class activation map (CAM)~\cite{zhou2016learning}, an weakly supervised approach for object localization, on top of the features extracted at different stages. CAM produces one heatmap for each class individually, and on each of these maps. We first rescale the maps to the original image size. If the maximal response across scales falls within one of the ground truth bounding boxes of the specified object, the predicted location is considered a valid localization.

Table~\ref{tab_localization} summarizes localization results. CHR based on DenseNet outperforms DenseNet by $5.61\%$ ($50.31\%$ vs $44.70\%$) for SIXray100 and $9.26\%$ ($43.87\%$ vs $34.61\%$) for SIXray1000.

\begin{figure}[t!]
\centering
\includegraphics[width=1\linewidth]{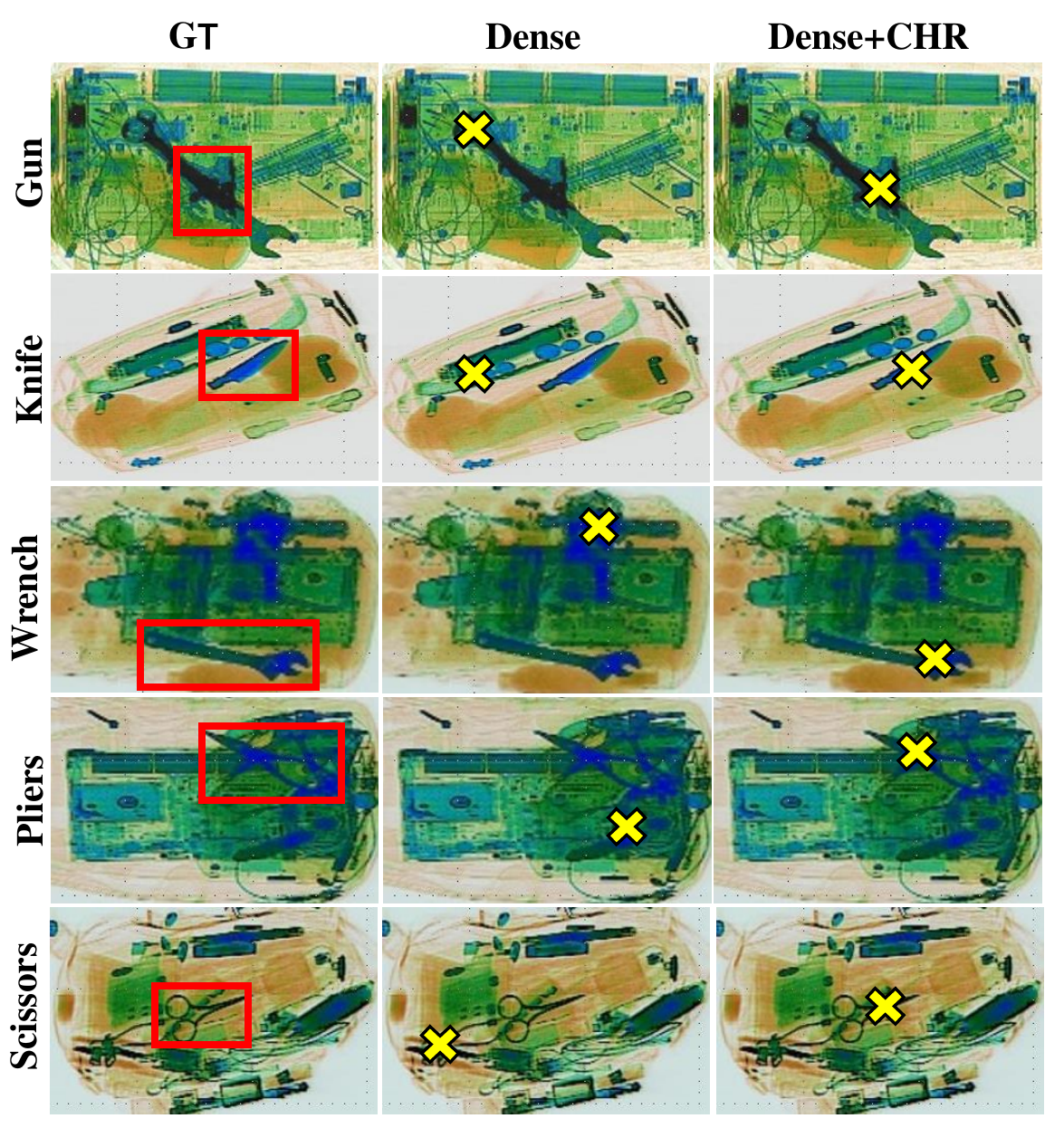}
\caption{Examples of object localization based on DenseNet, Which shows CHR is effective in complex background and overlapping images. (best viewed in color).}
\label{fig:localization}
\vspace{-0.2cm}
\end{figure}

Especially£¬ for {\em{Wrench}} of SIXray1000, Inception-v3+CHR outperforms Inception-v3 by $16.04\%$ ($23.53\%$ vs $7.49\%$). Again, we observe significant accuracy gain on deeper networks (which produces more powerful features) and larger negative-over-positive ratios. more localization results are shown in Figure~\ref{fig:localization}.

\subsection{Ablation Studies}
\label{sec:experiments:ablation}

In this part we provide diagnostic experiments.
These experiments are performed on all three subsets of SIXRay, which have different ratios of negative-positive samples.

First, we study the performance of hierarchical refinement -- the reversed connections, Table~\ref{tab:ablation}. It can be seen that the top-down refinement (ResNet34+HR) improves the classification and localization accuracies by $1\%$ and $6.52\%$ on SIXRay100, and $3.15\%$ and $2.13\%$ on SIXRay1000. (ResNet34+HR) outperforms the direct hierarchical fusion (ResNet34+H). The reason lies in that the information provided overlaps with regular networks, and the latter option provides more information to low-level features.

Second, we study the impact of different loss functions, Table~\ref{tab:ablation}. With the class-balance loss (ResNet34+CH), the classification and localization accuracies are improved by $1.00\%$ and $3.77\%$ on SIXRay100, and $3.10\%$ and $3.44\%$ on SIXRay1000. By combining hierarchical refinement with the class-balance loss (ResNet34+CHR), the classification and localization accuracies are improved by $2.37\%$ and $7.51\%$ on SIXRay100, and $5.48\%$ and $5.11\%$ over the baseline ResNet34, Table~\ref{tab:ablation}, which shows the significance of CHR on large-scale datasets with class imbalance.
\begin{table}[]
\small
\begin{center}
\setlength{\tabcolsep}{0.08cm}
\begin{tabular}{|l|c|c|c|c|c|c|}
\hline
Method & \multicolumn{2}{c|}{SIXray10} & \multicolumn{2}{c|}{SIXray100} & \multicolumn{2}{c|}{SIXray1000} \\
\hline\hline
ResNet34     & $74.83$ & $51.45$ & $52.74$ & $35.05$ & $33.26$ & $27.17$ \\
\hline
ResNet34+H   & $74.43$ & $49.91$ & $53.59$ & $38.70$ & $34.78$ & $28.68$ \\
\hline

ResNet34+CH  & $76.28$ & $48.01$ & $54.59$ & $42.47$ & $37.87$ & $32.12$ \\
\hline
ResNet34+HR  & $75.87$ & $50.19$ & $53.72$ & $41.57$ & $36.41$ & $29.30$ \\
\hline
ResNet34+CHR & $\mathbf{77.20}$ & $\mathbf{52.23}$ & $\mathbf{55.11}$ & $\mathbf{42.56}$ & $\mathbf{38.74}$ & $\mathbf{32.28}$ \\  \hline
\end{tabular}
\end{center}
\vspace{-0.1cm}
\caption{Classification mean Average Precision and localization accuracies ($\%$) on SIXRay subsets using options (refinement method, loss function, {\em etc.}) of CHR. The backbone is ResNet34. For the explanation of different options, see the main texts in Section~\ref{sec:experiments:ablation}.}
\label{tab:ablation}
\vspace{-0.3cm}
\end{table}

Note that the accuracy gain is achieved with a relatively small amount of extra computation. For example, ResNet34 requires $7.68\mathrm{ms}$ to process each testing image and ResNet34-CHR requires $8.28\mathrm{ms}$, both on a Tesla V100 GPU. Thus, $7.81\%$ extra time is used by CHR.

\subsection{ILSVRC2012 Classification}
\label{sec:experiments:ILSVRC2012}

Last but not least, we evaluate CHR on ILSVRC2012, a large-scale image classification dataset. This is to observe how CHR generalizes to natural image data, provided that it achieves significant accuracy gain on overlapping image data. ILSVRC2012 is a popular subset of the ImageNet databased, which has $1\rm{,}000$ classes and each of them contains a well-defined concept in WordNet. A total of $1.3\mathrm{M}$ training images and $50\mathrm{K}$ validation images are provided, both of which are roughly uniformly distributed over all classes.

We follow the standard training and testing pipelines, including the policies of model initialization, data augmentation, learning rate decay, {\em etc}. Since ILSVRC2012 is not an imbalanced dataset, we switch off the weight terms in the loss function which was designed for this purpose.

The top-$1$ error of CHR based on ResNet18 is $27.01\%$ \cite{he2016deep}, which slightly lower than the baseline by $0.87\%$ ($27.01\%$ vs $27.88\%$). Besides, the top-$1$ and top-$5$ errors of CHR based on ResNet50\cite{he2016deep} are $22.00\%$ and $6.22\%$. which are lower than the baseline by $0.85\%$ ($22.00\%$ vs $22.85\%$) and $0.49\%$ ($6.22\%$ vs $6.71\%$), respectively.
This slight but consistent accuracy gain delivers two-fold messages. The reversed connections in our approach which carries high-level supervision to mid-level features do not conflict with natural images -- although it aligns with overlapping image data much better. Given that the additional computational costs are almost negligible, it is worth investigating its extension in the natural image domains.

\section{Conclusions}
\label{sec:conclusions}

In this paper, we investigate prohibited item discovery in X-ray scanned images, which is a promising application in industry yet remains fewer studied in computer vision. To facilitate research in this field, we present SIXray, a large-scale dataset consisting of more than one million X-ray images, all of which were captured in real-world scenarios and therefore covered complicated scenarios. We manually annotated six types and more than $20\rm{,}000$ prohibited items, which is at least $100$ times larger than all existing datasets. In methodology, we formulate X-ray images as the overlap of several sub-images, therefore sampled from a mixture distribution. Motivated by filtering irrelevant information, we present an algorithm to refine mid-level features in a hierarchical and iterative manner. In practice, we switch off iteration to optimize the network weights in an approximate but efficient manner. A novel loss function is also built upon the hierarchical architecture to deal with heavy data imbalance between positive and negative classes. Beyond a few popular network backbones, our approach produces consistent gain in both classification and localization accuracy, establishing a strong baseline for the proposed task.

The future research mainly lies in two directions. First, the formulation of overlapping images from the penetration assumption is not accurate in many aspects -- we look forward to more effective approaches based on a better physical model. Second, the connection between overlapping images and natural images, {\em e.g.}, object occlusion, remains unclear -- studying this topic may imply some ways of extending these approaches to a wider range of applications.

{\small
\bibliographystyle{ieee}
\bibliography{egbib}
}
\end{document}